\documentclass{article}
\usepackage{spconf,amsmath,graphicx}
\usepackage[shortlabels]{enumitem}

\usepackage{ifthen}
\usepackage{fancyhdr}

\title{DUAL-TREE WAVELET SCATTERING NETWORK WITH PARAMETRIC LOG TRANSFORMATION FOR OBJECT CLASSIFICATION}
\name{Amarjot Singh and Nick Kingsbury}
\address{Signal Processing Group, Department of Engineering, University of Cambridge, U.K.}
\begin{document}

%
\maketitle
\begin{abstract}
We introduce a ScatterNet that uses a parametric log transformation with Dual-Tree complex wavelets to extract translation invariant representations from a multi-resolution image. The parametric transformation aids the OLS pruning algorithm by converting the skewed distributions into relatively mean-symmetric distributions while the Dual-Tree wavelets improve the computational efficiency of the network. The proposed network is shown to outperform Mallat's ScatterNet~\cite{Oyallon2015} on two image datasets, both for classification accuracy and computational efficiency. The advantages of the proposed network over other supervised and some unsupervised methods are also presented using experiments performed on different training dataset sizes.
\end{abstract}
\begin{keywords}
DTCWT, Scattering network, Convolutional neural network, Orthogonal least squares, CIFAR.
\end{keywords}
\section{Introduction}
\label{sec:intro}

Object classification is a difficult problem due to the translation, rotation and scale variability of objects within the images as well as external variabilities such as noise and illumination. Hand-engineered features such as SIFT~\cite{lowe} and HOG~\cite{dalal} modeled the geometric properties of the objects to achieve decent classification accuracy. However, these features have been recently replaced by trained networks~\cite{jia},~\cite{lee},~\cite{NIN}, especially, Convolutional Neural Networks (CNNs)~\cite{NIN} that have achieved state-of-the-art accuracy by learning invariant and discriminative class-specific image representations. Despite the success of CNNs, design and optimal configuration of these networks is not well understood which makes it difficult to develop these networks. 

Mallat~\cite{Jbruna2013},~\cite{sifre2013},~\cite{sifre2014},~\cite{Oyallon2015} has shown that ScatterNets incorporate geometric knowledge of images to produce discriminative and invariant (translation and rotation) representations which can give performance comparable to that of trained networks.~The invariants at the first layer of the network are obtained by filtering the image with multi-scale and multi-directional complex  Morlet wavelets followed by a point-wise nonlinearity and local smoothing. The high frequencies lost due to smoothing are recovered at the later layers using cascaded wavelet transformations, justifying the need for a multilayer network. A log transformation may be applied to de-correlate the multiplicative low-frequency components from the concatenated invariants obtained at all layers~\cite{Oyallon2015}. Next, orthogonal least squares (OLS) selects the subset of object class-specific dimensions across the training data, similar to that of the fully connected layers in CNNs~\cite{Oyallon2015}.The presence of outliers in the extracted features or unwanted features extracted from the background clutter, noise, and illumination can hinder feature selection due to their effect on the least squares parameter estimates. Hence, it is important to introduce approximate symmetry in the extracted features to suppress the effect of these outliers. 

We propose an improved computationally efficient ScatterNet that extracts relatively symmetric translation invariant representations from a multi-resolution image using the \textit{dual-tree complex wavelet transform} (DTCWT)~\cite{Kingsbury1998} and the proposed parametric log transformation layer. Here, we only introduce translation invariance, as the orientation of an object in the image plane is often well-known as a strong prior (e.g. side-view images). The OLS layer next selects a subset of object specific components without undesired bias from outliers due to the introduced symmetry. The selected features are finally used by a Gaussian-kernel support vector machine (G-SVM) to perform object classification on CIFAR-10 and CIFAR-100 datasets.  

 \begin{figure*}[t!] 
\centering    
\includegraphics[width = 0.91\textwidth, height = 9.7 cm]{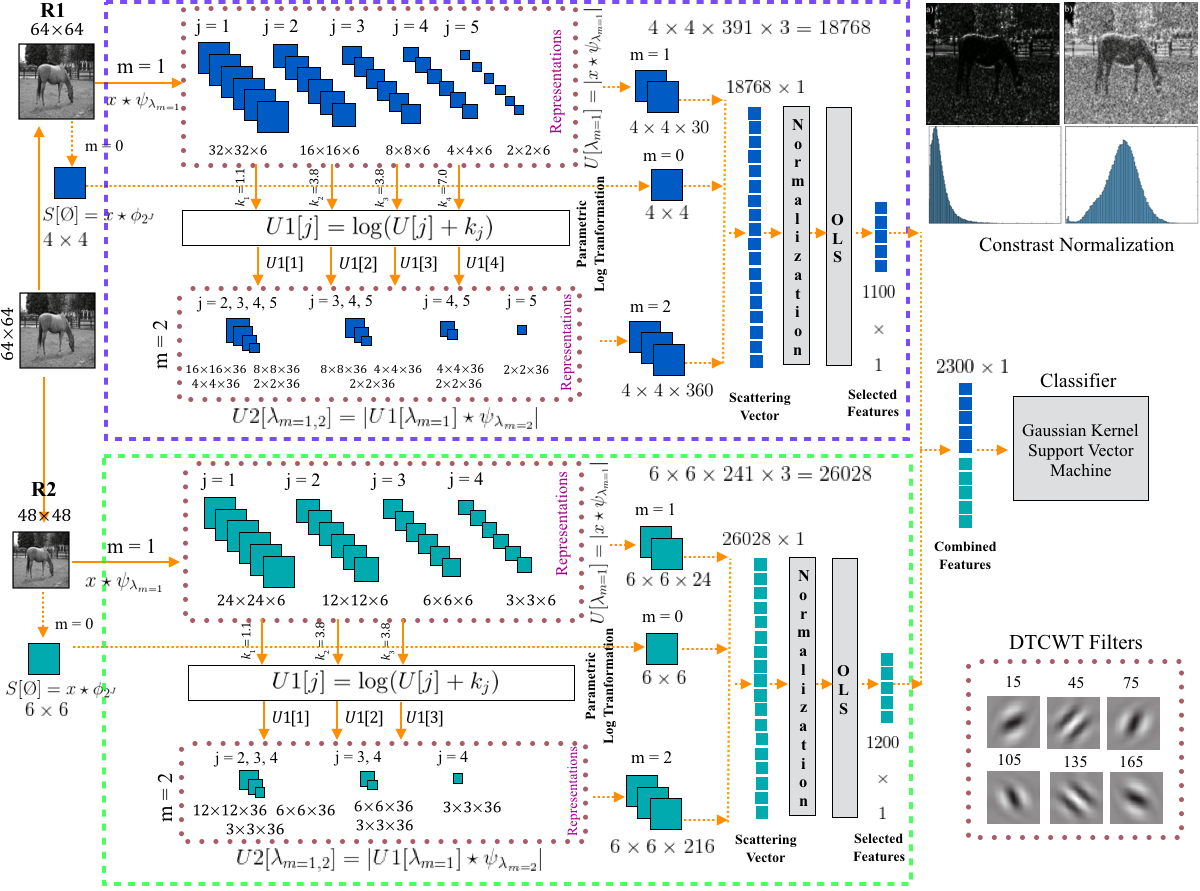}
\caption[Scattering operations performed on a signal to extract filtered response with translation invariance and to recover the lost frequency components. ]{\small{{Illustration shows the input image ($x$) of size $64 \times 64$ resized to images of resolution, R1 ($64 \times 64$ ($x$)) and R2 ($48 \times 48$ ($x_{1}$) respectively. Image representations at m = 1 are obtained using DTCWT filters at 5 scales for R1, 4 scales for R2  and 6 orientations ($x \star \psi_{\lambda_{m = 1}}$). Next, L2 non-linearity (complex modulus) is applied on the representations to obtain the regular envelope $|x\star \psi_{\lambda_{m = 1} }|$. Log transformation $U1[j] = \log(U[j] + k_{j})$ with parameters $k_{j}$ is applied on the envelope for all scales $j$ except the coarsest scale. Next, local smoothing is applied to extract the translation invariant coefficients $U1[\lambda_{m = 1}] \star \phi_{2^J}$. The information lost due to smoothing are recovered by cascaded wavelet filtering at the second layer $|U1[\lambda_{m = 1}] \star \psi_{\lambda_{m = 2}}|$. Translation invariance is introduced in the recovered frequencies using L2 non-linearity and local smoothing $U2[\lambda_{m = 1},\lambda_{m = 2}] \star \phi_{2^J}$. The contrast normalization effect of the parametric log transformation is shown in the top right while the DTCWT filters at six fixed orientations are shown in the bottom.}}}
\label{fig:scatter00}
\end{figure*} 
  
The contributions of the paper are as follows:

\begin{itemize}[topsep=2.2pt]
\itemsep0em 
\item \textit{Multi-resolution Input Image}: The input image is transformed into multi-resolution images of 2 or more different sizes such that the dual-tree wavelet decompositions produce more densely spaced feature maps over scale. These allow the OLS algorithm to learn additional discriminatory features which can aid the classification.  
 
\item \textit{Parametric Log transformation}: Log transformation reduces the effect of outliers by introducing approximate symmetry in representations with parameters learnt from the data. The transformation also de-correlates the multiplicative low-frequency components (illumination) while simultaneously creating a form of contrast normalization which enhances weaker features. 

\item \textit{Computational Efficiency}: 
Dual-tree wavelets are used as opposed to Morlet ~\cite{Jbruna2013} because of their discrete form, short support, perfect reconstruction, and limited redundancy~\cite{Kingsbury1998}. They provide similar rich features to Morlet wavelets but with less computation and somewhat lower redundancy in the output vectors. In addition, dual-tree wavelets can be efficiently implemented in the spatial domain, rather than requiring the complexities and constraints of Fourier domain filtering. 

  \end{itemize}

The proposed network improves on Mallat's ScatterNet on classification accuracy and computational efficiency on two datasets. Multiple experiments on different training dataset sizes are performed to highlight the advantages of the proposed network against supervised and unsupervised methods.

  The paper is divided into the following sections. Section 2 briefly presents our proposed DTCWT scattering network with parametric log transformation. Section 3 presents the experimental results while Section 4 draws conclusions.

\section{DTCWT ScatterNet}
\label{sec:typestyle}
The proposed ScatterNet uses dual-tree wavelets to decompose the multi-resolution input image into multi-scale and multi-directional representations at multiple layers, because the DT-CWT is lossless and has high computational efficiency in the spatial domain. The parametric log transformation is applied to the outputs of the first scatternet layer to introduce relative symmetry to the distributions of coefficient magnitudes and thus aid OLS feature selection. Subsequent scatternet layers then apply local smoothing and bandpass filtering with wavelet-modulus operations to gradually increase translation invariance while preserving information about the higher frequency components of the image~\cite{Jbruna2013}. Below we present the formulation of the proposed ScatterNet for a single input image which may then be applied to each of the multi-resolution images.

An input image $x$ is filtered using dual-tree complex wavelets $ x\star \psi_{\lambda_{1} }$ where $\lambda_1 = (j,r)$.  At the first layer, the real and imaginary parts of the complex coefficients are combined from real filters in the dual tree using:
\vspace{-0.3em}
\begin{align}
 x\star \psi_{\lambda_{1} } = x\star\psi_{\lambda_{1} }^{a} + \iota x\star\psi_{\lambda_{1} }^{b}  
\end{align}
where $\psi^{a}$ is the real and $\psi^{b}$ the imaginary part of the wavelet. The six orientations (r) in the transform are pre-defined to be: $15^\circ, 45^\circ, 75^\circ, 105^\circ, 135^\circ$ and $165^\circ$. 

The wavelet filtering signal commutes with translations, and is therefore not translation invariant. To build a more translation invariant representation, a point-wise $L_{2}$ non-linearity is applied to the filtered signal, as described below:
\vspace{-0.3em}
\begin{equation}
U[\lambda_{m = 1}] = |x\star \psi_{\lambda_{1} }| = \sqrt{|x\star \psi_{\lambda_{1} }^{a}|^2 + |x\star \psi_{\lambda_{1} }^{b}|^2} 
\end{equation}
This step produces the regular envelope of the filtered signal and reduces the redundancy of each representation to 2:1. $L_{2}$ is a good non-linearity as it is stable to deformations and additive noise~\cite{Jbruna2013}. However, the representations may also contain outliers that can hinder the performance of the orthogonal least squares based feature selection layer (explained in Section. 1). Hence, the parametric log transformation layer is applied to all the oriented representations ($U[j]$) extracted at a particular scale $j$ with a parameter $k_{j}$, to reduce the effect of outliers by introducing relative symmetry, as shown below: 
 \vspace{-0.3em}  
   \begin{equation}
  U1[j] = \log(U[j] + k_{j}), \quad U[j] = |x\star \psi_{j}|, 
\end{equation}
Good symmetry is achieved for the distribution of oriented representations obtained by selecting the parameter $k_{j}$ that minimizes the difference between the mean and median of the distribution. The parametric log transformation also de-correlates the low-frequency multiplicative components arising due to illumination variation and noise~\cite{Oyallon2015} as well as \textit{normalizing the contrast} of the representations by elevating the weak features and suppressing the stronger ones as shown top right corner in Fig. 1. 

Next, a local average is computed on the envelope $|U1[\lambda_{m = 1}]|$ that aggregates the coefficients to build the desired translation-invariant representation: 
\vspace{-0.3em}
\begin{equation}
S[\lambda_{m = 1}] = |U1[\lambda_{m = 1}]| \star \phi_{2^J}
\end{equation}

The high frequency components lost due to smoothing are retrieved by cascaded wavelet filtering performed at the second layer. The retrieved components are again not translation invariant. Translation invariance is achieved by first applying the L2 non-linearity of eq(2) to obtain the regular envelope:
\vspace{-0.3em}
\begin{equation}
U2[\lambda_{m = 1},\lambda_{m = 2}] = |U1[\lambda_{m = 1}] \star \psi_{\lambda_{m = 2}}|
\end{equation}

A local-smoothing operator is then applied to the regular envelope ($U2[\lambda_{m = 1},\lambda_{m = 2}]$) to extract the desired second layer ($m = 2$) translation invariant coefficients: 
\vspace{-0.3em}
\begin{equation}
S[\lambda_{m = 1},\lambda_{m = 2}] = U2[\lambda_{m = 1},\lambda_{m = 2}] \star \phi_{2^J}
\end{equation}
The scattering coefficients obtained at each layer are:
\vspace{-0.3em}
\begin{equation}
S = \begin{pmatrix}
x \star \phi_{2^J} \\
U1[\lambda_{m = 1}] \star \phi_{2^J}
\\ U2[\lambda_{m = 1},\lambda_{m = 2}] \star \phi_{2^J}
\end{pmatrix}_{j = (2,3,4,5...)}
\end{equation} 
The coefficients extracted from each layer are concatenated to generate a feature vector for each of the images in the training dataset as shown in Fig. 1. The scattering feature vectors are then normalized across each dimension and given as input to the feature selection layer. 

The feature selection layer is implemented using a supervised orthogonal least square (OLS) regression~\cite{Blumensath} that greedily selects discriminative features specific to class $C$ with a one-versus-all linear regression using the following indicator function:
\vspace{-0.3em}
\begin{equation}
\begin{aligned}
f_{C}(x) = \left\{\begin{matrix}
1 \quad $\text{if x belongs to class C}$ \\ 
0 \quad $\text{otherwise}$ \quad \quad \quad \quad \quad 
\end{matrix}\right.
\end{aligned}
\end{equation}
The regression is applied to a training set of scattering feature vectors where each vector of $N$ dimensions is reduced to $N'$ selected dimensions ($N' << N$) that belong to a specific class $C$. Let $(\Phi^{M\times N}_{t})_{C}$ be the dictionary at the $t^{th}$ iteration for a specific class $C$. The $t^{th}$ feature $x$ is selected such that the linear regression of $f_{C}(x)$ has a minimum mean-squared error, computed on the training set corresponding to class $C$. The reduced training feature dataset is given as input to the G-SVM that learns the weights that best discriminate the classes in the dataset. Feature selection makes training and applying a classifier more efficient due to the decreased vector size. It also tends to improve performance by eliminating unnecessary components of the input and their associated noise.

\section{Overview of Results}
\label{headings}
The performance of the proposed network is evaluated on CIFAR-10 and CIFAR-100 datasets with 10 and 100 classes respectively. Each dataset contains a total of 50000 training and 10000 test images of size $32\times32$ equally divided between the classes. The evaluation is performed on the classification accuracy, computational efficiency and feature richness. A comparison with Mallat's ScatterNet~\cite{Oyallon2015}, unsupervised~\cite{sohn},~\cite{jia} and supervised methods~\cite{NIN} is also performed.

In order to extract the scattering representations, every 32$\times$32 image is first upsampled into two images of resolution 64$\times$64 (R1) and 48$\times$48 (R2). The upsampled image is then transformed into two images of resolution 64$\times$64 (R1) and 48$\times$48 (R2). R1 and R2 are decomposed using DTCWT filters with 6 fixed orientations at 5 and 4 scales respectively, followed by L2 non-linearity, as shown in Fig. 1. Next, the log transformation is applied to the representations (except the ones obtained at the coarsest scale) obtained from both R1 and R2 pipeline with parameters $k_{1}$ = 1.1, $k_{2}$ =3.8, $k_{3}$ =3.8 and $k_{4}$ =7 chosen for scale j = 1, 2, 3 and 4 respectively. An smoothing operator is then applied to introduce translation invariance in the representations. The classification accuracy for representations obtained at various scales (J), with and without the use of parametric log transformation and the concatenated coefficients at m=1 with G-SVM, are shown for both R1 and R2 pipelines in Table 1. The G-SVM parameter (c) is selected as 14 while gamma parameter is set to 0.00002 using 5-fold cross validation on the training feature set. We see that the parametric log transformation results in a small improvement in classification accuracy. 

The information lost due to smoothing at the first layer is retrieved at the next layer using cascaded filtering as shown in Fig. 1. The retrieved information is made translation invariant by local smoothing. Representations for the three color channels at m = 0, 1, 2 are concatenated to produce a 18768 (6256 $\times$ 3) dimensional vector for R1 image and a vector of length 26028 (8676 $\times$ 3) for R2 as shown in Fig. 1. OLS is then applied on the training dataset ($50000 \times 18768$) to select 108 dimensions per class resulting into a total of 1080 discriminative dimensions for every R1 image ( $50000 \times 1080$).  Similarly, 1200 dimensions per image are chosen for the R2 image. This reduced feature dataset results in a classification accuracy of 81.6\% (80.7\% without log transformation) for R1 images while an accuracy of 81.8\% (80.9\% without log transformation) is recorded for R2 images, using the above-mentioned SVM for the CIFAR-10 datasets as shown in Table. 1. A classification accuracy of 82.4\% is obtained by concatenating the selected dimensions of R1 and R2. A decrease in classification accuracy is recorded on selecting more than the above-mentioned feature dimensions. \vspace{-5mm}

\begin{table}[!h]
\centering
\caption{\small{Accuracy (\%) on CIFAR-10 for both R1 and R2 for each scale (J) and coefficients at m = 1, with and without applying log transformation. The accuracy for features selected from the final scattering vector at $m_{1,2}$ using OLS is presented in the last column.}}
\label{components}
\begin{tabular}{c|ccccccc}
\hline
 & \small{$J=1$} & \small{$J=2$} & \small{$J=3$} & \small{$J=4$} & \small{$m_{1}$} & \small{$m_{1,2}$} \\ \hline \hline
 \small{R1: No-log} & 62.7 & 66.9 & 69.0 & 70.2 & 70.4 & 80.7  \\
 \small{R1: log } & \textbf{65.6} & \textbf{69.9} & \textbf{71.5} & \textbf{72.4} & \textbf{72.5}  & \textbf{81.6} \\ \hline
 
  \small{R2: No-log} & 65.9 & 70.0 & 71.2 & -- & 71.7 & 80.9  \\
 \small{R2: log} & \textbf{68.0} & \textbf{71.5} & \textbf{72.6} & -- & \textbf{73.4}  & \textbf{81.8} \\

\end{tabular}
\end{table}
\vspace{-3mm}

Next, scattering coefficients extracted using DTCWT ScatterNet with the above-mentioned parameters result in a classification accuracy of 56.7\% for the CIFAR-100 dataset, as shown in Table. 2. The translation invariant coefficients extracted using the proposed network outperform the translation as well as Roto-translation invariant features of Mallat's ScatterNet~\cite{Oyallon2015}, on both datasets. The network also outperformed state-of-the-art unsupervised methods~\cite{sohn},~\cite{jia} but underperformed by nearly 10\% against supervised deep learning models~\cite{NIN}, as shown in Table. 2. \vspace{-5mm}

\begin{table}[!h]
\centering
\caption{\small{Accuracy (\%) and comparison on both datasets. Pro.: Proposed, Sup: Supervised and Unsup: Unsupervised, learning.}}
\label{MNIST}
\begin{tabular}{c|cccc}
\hline
 Dataset & Pro. & ScatNet~\cite{Oyallon2015} & Unsup & Sup\\
 \hline
 CIFAR-10 & \textbf{82.4} & 81.6 & 82.2~\cite{sohn} & 89.6~\cite{NIN}\\
 CIFAR-100 & \textbf{56.7} & 55.8 & 54.2~\cite{jia} & 64.3~\cite{NIN}\\

\end{tabular}
\end{table}
\vspace{-3mm}

The proposed network can be an attractive choice over Mallat's ScatterNet due to its computational efficiency and gain in classification accuracy. The proposed network extracts the coefficients from both R1 and R2 images in almost three-quarters (0.78 (s)) of the time needed by Mallat's network (0.98 (s)) to decompose only the R1 image, as shown in Table. 3. This marginal difference is significant for large image datasets such as CIFAR. In addition, since the scattering vector produced by the proposed network is smaller (44796) as compared to Mallat's network (113712) (three-layer network)~\cite{Oyallon2015}, the OLS layer can select the desired feature dimensions (1080 and 1200) in almost 3/4 of the time (2.21(h) vs 3.22(h)). The selected dimensions with OLS from the scattering vector are more for the proposed network (1080 + 1200 = 2300) as compared to Mallat's network (1080). This suggests that the features extracted by the proposed network are significantly richer in information as compared to Mallat's network as feature richness is defined as the number of dimensions selected with OLS divided by the total feature dimensions. The simulations are computed on a server with 32 Gb RAM per node in uniform conditions. \vspace{-5mm}

\begin{table}[!h]
\centering
\caption{\small{Arc.: Architectures, Pro.: Proposed, R1, R2: Resolution - 1,2  pipeline, FVL: Feature vector length, SD: Selected dimensions using OLS, FR: Feature richness (\%), TS (s): Scattering time an image in seconds, T-OLS: Feature selection time using OLS in hours.}}
\label{MNIST}
\begin{tabular}{c|ccccc}
\hline
\small{Arch.} & \small{FVL} & \small{SD} & \small{FR (\%)} & \small{TS (s)} & \small{T-OLS (h)}\\
 \hline
\small{ScatNet~\cite{Oyallon2015}}  & \small{113712} & \small{2000} & \small{1.75} & \small{0.98} & \small{3.22}\\
\small{R1} & \small{18762} & \small{1100} & \textbf{\small{5.86}} & \textbf{\small{0.46}} & \textbf{\small{1.07}}\\
\small{R2} & \small{26028} & \small{1200} & \textbf{\small{4.61}} & \textbf{\small{0.32}} & \textbf{\small{1.14}}\\ \hline
\small{Pro. (R1+R2)} & \small{44796} & \small{2300} & \textbf{\small{5.13}} & \textbf{\small{0.78}} & \textbf{\small{2.21}}\\
\end{tabular}
\end{table}
\vspace{-3mm}

However, supervised models require large training datasets to learn which may not exist for most application. Table. 4 shows that DTCWT ScatterNet outperformed LeNet~\cite{lee} and Network in Network (NIN)~\cite{NIN} supervised learning networks on the CIFAR-10 datasets with less than 10k images. The experiments were performed by dividing the training dataset of 50000 images into 8 datasets of different sizes. The images for each dataset are obtained by randomly selecting the required number of images from the full 50000 training dataset. It is made sure that an equal number of images per object class are sampled from the training dataset. The full test set of 10000 images is used for all the experiments. Deeper models like NIN~\cite{NIN} result in low classification accuracy due to their inability to train on the small training dataset.

\vspace{-5mm}
\begin{table}[!h]
\centering
\caption{\small{Comparison of Proposed (Pro.) network on accuracy (\%) with two supervised learning methods (LeNet~\cite{lee} and NIN: Network in Network~\cite{NIN} against different training dataset sizes on CIFAR-10.}}
\label{MNIST}

\begin{tabular}{c|cccccccc}
\hline
\small{Arch.} & \small{300} & \small{500} & \small{1K} & \small{2K} & \small{5K} & \small{10K} & \small{20K} & \small{50K}\\
 \hline
\small{Pro.} & \small{\textbf{39.3}} & \small{\textbf{48.8}} & \small{\textbf{55.9}} & \textbf{\small{61.8}} & \small{\textbf{67.0}} & \small{\textbf{72.9}} & \small{{76.8}} & \small{82.4}\\
\small{LN} & \small{{34.9}} & \small{44.7} & \small{53.1} & \small{57.9} & \small{{63.0}} & \small{69.0} & \small{74.0} & \small{77.6}\\
\small{NIN} & \small{{10.1}} & \small{10.3} & \small{10.9} & \small{40.4} & \small{{63.4}} & \small{72.0} & \textbf{\small{83.1}} & \textbf{\small{89.6}}\\
\end{tabular}
\end{table}
\vspace{-8mm}

\section{Conclusion}
The paper proposes an improved version of Mallat's ScatterNet using   dual-tree wavelets and parametric log non-linearity. The DTCWT ScatterNet gives enhanced performance on classification accuracy and computational efficiency as compared to Mallat's ScatterNet on two datasets. The network has also shown to outperform unsupervised learning methods while evidence of the advantage of DTCWT ScatterNet over supervised learning (CNNs) methods is presented for applications with small training datasets.  
\bibliographystyle{IEEEbib}
\bibliography{refs}

\end{document}